\pdfoutput=1

\documentclass[journal]{IEEEtran}
\usepackage{cite}
\usepackage{times}
\usepackage{soul}
\usepackage{url}
\usepackage[hidelinks]{hyperref}
\usepackage[utf8]{inputenc}
\usepackage[small]{caption}
\usepackage{graphicx}
\usepackage{amsmath}
\usepackage{amsthm}
\usepackage{amstext}
\usepackage{amssymb}
\usepackage{booktabs}
\usepackage{float}
\usepackage{xcolor}
\usepackage{tabularx}
\usepackage{multirow}
\usepackage{makecell}
\usepackage{bm}
\usepackage{algorithm}
\usepackage{algorithmic}
\usepackage[footnotesize]{subfigure}

\usepackage{colortbl}

\newcommand{\tabitem}{~~\llap{\textbullet}~~}

\definecolor{intro_green}{RGB}{56, 87, 35}
\definecolor{intro_purple}{RGB}{112, 48, 160}
\definecolor{intro_orange}{RGB}{197, 90, 17}

\usepackage{tikz}
\newcommand*\circled[1]{\tikz[baseline=(char.base)]{
            \node[shape=circle,draw,inner sep=0.4pt] (char) {#1};}}
            
\usepackage{newfloat}
\usepackage{listings}
\floatstyle{ruled}
\newfloat{listing}{tb}{lst}{}
\floatname{listing}{Listing}

\ifCLASSINFOpdf
\else
\fi

\begin{document}
%
% paper title
% Titles are generally capitalized except for words such as a, an, and, as,
% at, but, by, for, in, nor, of, on, or, the, to and up, which are usually
% not capitalized unless they are the first or last word of the title.
% Linebreaks \\ can be used within to get better formatting as desired.
% Do not put math or special symbols in the title.
\title{Controllable Dialogue Generation with Disentangled Multi-grained Style Specification and Attribute Consistency Reward}
%
%
% author names and IEEE memberships
% note positions of commas and nonbreaking spaces ( ~ ) LaTeX will not break
% a structure at a ~ so this keeps an author's name from being broken across
% two lines.
% use \thanks{} to gain access to the first footnote area
% a separate \thanks must be used for each paragraph as LaTeX2e's \thanks
% was not built to handle multiple paragraphs
%

\author{{Zhe Hu}$^{1*}$,  {Zhiwei Cao}$^{2*}$, {Hou Pong Chan}$^{3}$,  {Jiachen Liu}$^{1}$, {Xinyan Xiao}$^{1}$, {Jinsong Su}$^{2 \dagger}$, and {Hua Wu}$^{1}$
\\
  $^{1}$Baidu Inc 
  $^{2}$Xiamen University
  $^{3}$ University of Macau
 \\
  $^{1}${\tt \{huzhe01,liujiachen,xiaoxinyan,wu\_hua\}@baidu.com} \\
  $^{2}${\tt lines@stu.xmu.edu.cn,jssu@xmu.edu.cn}, 
  $^{3}${\tt hpchan@um.edu.mo}
\thanks{* These authors contributed equally. Work done while Zhiwei Cao was an intern at Baidu}
\thanks{$\dagger$ Corresponding author}
  }

\maketitle

% As a general rule, do not put math, special symbols or citations
% in the abstract or keywords.
\begin{abstract}
Controllable text generation is an appealing but challenging task, which allows users to specify particular attributes of the generated outputs.
In this paper, we propose a controllable dialogue generation model to steer response generation under multi-attribute constraints. 
Specifically, we define
and categorize the commonly-used control attributes into global and local ones,
which possess different granularities of effects on response generation.
Then, we significantly extend the conventional seq2seq framework by introducing a novel two-stage decoder, which first uses a \textit{multi-grained style specification layer} to impose the stylistic constraints and determine word-level control states of responses based on the attributes,
and then employs a \textit{response generation layer} to generate final responses maintaining both semantic relevancy to the contexts and fidelity to the attributes. 
Furthermore, we train our model with an attribute consistency reward to promote response control with explicit supervision signals. 
Extensive experiments and in-depth analyses on two datasets indicate that our model can significantly outperform competitive baselines in terms of response quality, content diversity and controllability.
\end{abstract}

% Note that keywords are not normally used for peerreview papers.

% For peer review papers, you can put extra information on the cover
% page as needed:
% \ifCLASSOPTIONpeerreview
% \begin{center} \bfseries EDICS Category: 3-BBND \end{center}
% \fi
%
% For peerreview papers, this IEEEtran command inserts a page break and
% creates the second title. It will be ignored for other modes.
\IEEEpeerreviewmaketitle

\section{Introduction}
\IEEEPARstart{A}{s} a long-standing task in natural language processing, dialogue generation aims to automatically produce responses given input contexts.
In this aspect, the dominant methods are neural sequence-to-sequence (seq2seq) models \cite{cho-etal-2014-learning,vinyals2015neural,shang-etal-2015-neural} trained to maximize the log-likelihood over responses in an end-to-end fashion.
However, such generated responses not only lack controllability and interpretability~\cite{hua-wang-2019-sentence}, but also tend to be boring with genericness and repetitiveness~\cite{Li2016naacl}.

\begin{figure}[t]
    \centering
    \includegraphics[scale=0.53]{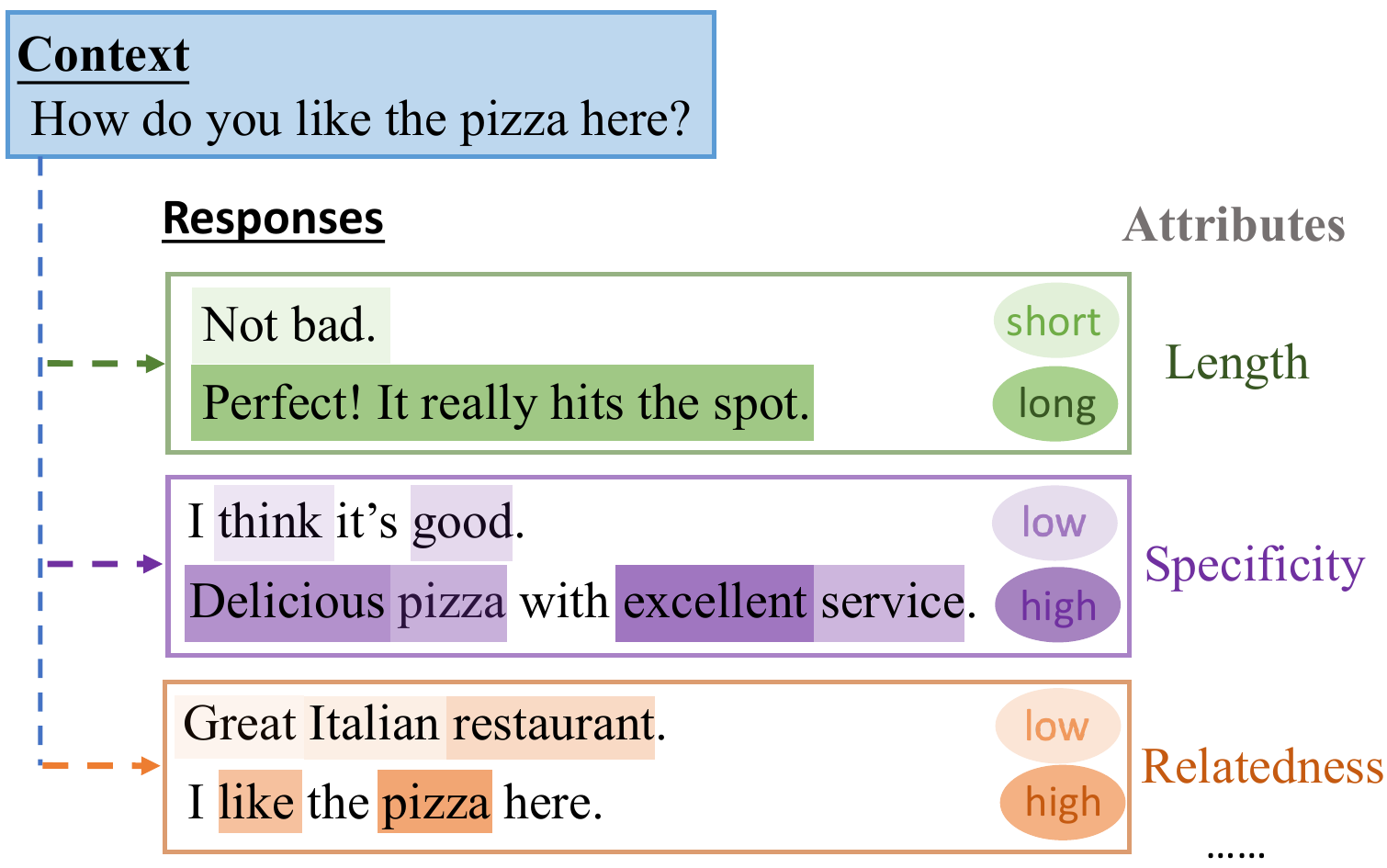}
    \captionof{figure}{ 
    There exist multiple responses with different control attributes for the same input context. \textcolor{intro_green}{Length} controls response from a global perspective, whereas 
    \textcolor{intro_purple}{Specificity} and \textcolor{intro_orange}{Relatedness} can be directly reflected on each response token. Color brightness indicates the corresponding attribute values.
    }
    \vspace{-2mm}
    \label{fig:sample_intro}
\end{figure}

One important reason 
for these defects stems from the fact that the above models neglect the one-to-many relationship between context and response~\cite{zhang-etal-2018-learning-control,xu-etal-2019-neural}. As shown in Figure \ref{fig:sample_intro},
for the same context, 
there exist multiple valid responses corresponding to different attributes.
Generally, the conventional methods maximizing the likelihood of responses given contexts are unable to explicitly learn the correspondence relationship between response and attributes.
Thus, incorporating explicit control into the generation is crucial to tackle the above defects.

To achieve this goal, many efforts have been devoted to exploring control variables for
dialogue generation~\cite{zhao-etal-2017-learning,see-etal-2019-makes,zhang-etal-2018-learning-control,zhou-wang-2018-mojitalk,ke-etal-2018-generating}. However, these studies mainly focus on leveraging a single attribute to control a specific aspect, which is unsuitable for real applications that involve multiple attributes. 
Recently, \cite{xu-etal-2019-neural} propose a memory-enhanced seq2seq model to govern response generation with multiple variables,
% \textcolor{blue}{ ## annotation: 2022-0511 ##
and \cite{russo-etal-2020-control} 
employ adversarial learning to generate sentences controlling multiple semantic
and syntactic attributes.
% }
Nevertheless, there still exist three drawbacks in these studies:
1) They equally consider all attributes, but as shown in Figure~\ref{fig:sample_intro}, different attributes impact generation with varying effects (e.g., some attributes control response globally whereas some attributes possess the fine-grained influence on each response token). This hinders the model flexibility to accurately reflect the attributes on outputs. 2) Controllable generation involves a complicated disentanglement process, where the model is required to generate responses maintaining both relevancy to the contexts and fidelity to the attributes, especially under the multi-attribute constraints. 
However, existing dialogue models couple style specification and response generation altogether in a single module, which leads to low interpretability and controllability. 
3) Current methods are usually trained with the maximum likelihood objective and only learn weak connections between the control attributes and responses, thus often generating outputs inconformable to the attributes. 

In this paper, we propose \textbf{CRAYON}, a framework to generate \underline{C}ontrollable \underline{R}esponse with multi-gr\underline{A}ined st\underline{Y}le specification and attribute c\underline{ON}sistency reward.
% To address the first drawback, 
We consider important dialogue attributes including \emph{specificity}, \emph{sentiment}, \emph{response-relatedness}, \emph{question-asking} and \emph{response length}. We further classify these attributes into two categories based on their properties: \emph{global attributes} affecting the generation of responses from an overall perspective, 
and \emph{local attributes} influencing the generation of each response word.
Such classification enables our model to more flexibly and accurately control response generation at different levels. 

To tackle the second drawback, we separate the control states and semantic states by dividing the generation process into two steps of style specification and surface generation, which further improves the model controllability and interpretability.
Specifically, as a significant extension of conventional seq2seq method~\cite{cho-etal-2014-learning},
our model is equipped with a novel two-stage controlled decoder: 1) a \textbf{multi-grained style specification layer} first imposes \textit{stylistic constraints} and generates a sequence of word-level control states based on the attributes, and 2)
a \textbf{response generation layer} then handles \textit{semantic requirements} on relevancy and produces a final response. To the best of our knowledge, our work is the first attempt that applies word-level style specification with multi-grained control to achieve better disentanglement for controllable generation.
 
Furthermore, we apply reinforcement learning (RL) with Markov Decision Process to optimize the model towards dedicated reward functions. During this process, we design reward functions that explicitly encourage the generated responses to satisfy the attribute constraints. By introducing direct supervision signals on attribute fidelity, our model is able to generate more diverse responses with better controllability. 

We carry out experiments on two dialogue generation datasets, Persona-Chat and DailyDialog. Automatic and human evaluations show that our model significantly outperforms both controllable and non-controllable baselines towards response quality, content diversity and controllability, demonstrating the ability to disentangle the complex controllable generation under multi-attribute constraints.

\section{Related Work}
\noindent\textbf{Dialogue Generation.} 
Neural response generation models are mostly based on seq2seq framework~\cite{sutskever2014sequence,Li2016naacl}. To improve the quality of response and address problems such as generic and safe response~\cite{sutskever2014sequence,Mou2016coling}, many extensions under the encoder-decoder framework have been proposed. For instance, maximum mutual information objective~\cite{Li2016naacl} or diverse beam search~\cite{Mou2016coling} are utilized to address the generic response issue during decoding.
Besides, some work also tackles this problem during model training. For example, adversarial learning~\cite{Li2017emnlp,Zhang2018nips} or reinforcement learning~\cite{Li2016emnlp-rf,yao2016attentional,xu2018towards,saleh2020hierarchical} based methods could directly improve the quality of responses. Some studies also adopt latent
variables to capture the response variation and
control the generation~\cite{zhao-etal-2017-learning,gao2019generating}, yet these latent variables are difficult to
explain and hard to control the generation with specific attributes. \cite{sankar-ravi-2019-deep} leverage discrete attributes with reinforcement learning to promote response diversity. However, RL is only used for dialogue attribute prediction without direct supervision on responses.

\begin{figure}[t]
    \centering
    \includegraphics[scale=0.53]{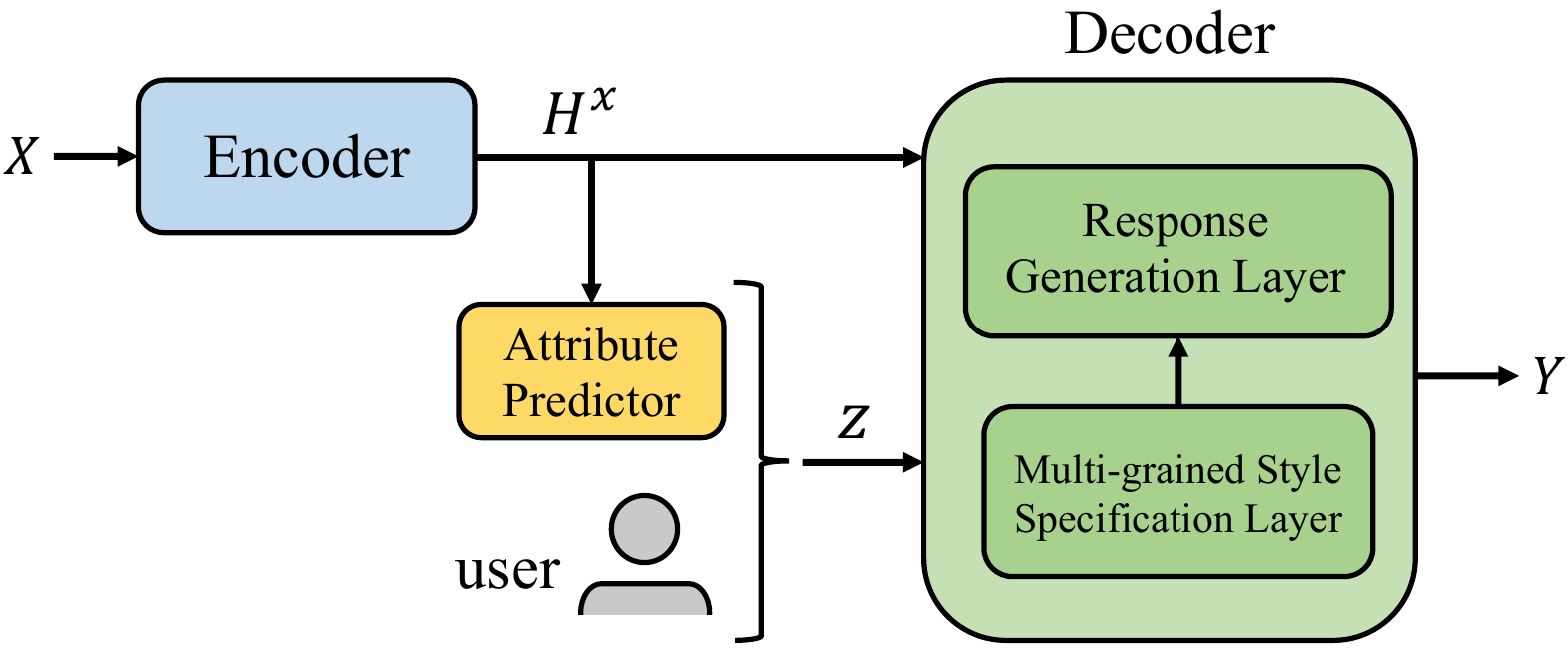}
    \vspace{2mm}
    \captionof{figure}{ 
    The overview of \textsc{CRAYON}. $\bm{x}$, $\bm y$ and $\bm z$ 
    % \textcolor{blue}{ ## annotation: 2022-0511 ##
    denote
    % } 
    the input context, response and control attributes respectively. The attributes can be either provided by the user or automatically inferred from the contexts.
    }
    % \vspace{-4mm}
    \label{fig:overall_structure}
\end{figure}

\smallskip
\noindent\textbf{Controllable Generation.}
Our work is also in line with controllable generation.
Recent work incorporates control attributes such as specificity~\cite{takayama-arase-2020-consistent}, topic~\cite{Baheti2018emnlp-topic}, dialogue acts~\cite{xu2018towards}, phrase~\cite{wu2020controllable,chan-etal-2021-controllable}, and style~\cite{Wang2017emnlp-style,gao-etal-2019-structuring,zhou-etal-2020-exploring}.
 \cite{hedayatnia-etal-2020-policy} use a dialogue policy to control responses at the turn and sentence
levels with grounded knowledge. \cite{xu-etal-2019-neural} propose a memory-enhanced seq2seq model with a multi-attribute controlling mechanism. \cite{gupta2020controlling} control dialogue generation based on the semantic frames of retrieved exemplars to improve coherency.

% \textcolor{blue}{ ## annotation: 2022-0511 ##
Conditional training~\cite{keskar2019ctrl} and weighted decoding~\cite{see-etal-2019-makes} are commonly used for controllable generation.
Some  works also control the generation process by modifying the attribute-related word distributions~\cite{ke-etal-2018-generating,wang-etal-2018-learning-ask},
which estimate a type distribution over word types during generation process. However, their word type estimation aims to modulate the final output distributions, and does not address the disentanglement problem of multiple attribute constraints.
Compared with the previous work~\cite{zhou-wang-2018-mojitalk} that adopts RL training to control the emotion of the generated outputs, our RL reward consists of multiple attribute constraints to jointly steer the generation process, which is more difficult. We further address the disentanglement problem of multi-attribute controlled generation with a separate style specification layer.
In conclusion, our work is different from the above methods in the following aspects: 1) We divide control attributes into global and local ones, so as to facilitate more flexible controls in different granularities; 
2) We adopt multi-grained style specification and response generation to address complicated disentanglement by separating the control states and semantic states. Compared with the methods estimating word type distributions, our multi-grained style specification layer imposes the controlled attributes into an appropriate local attribute state sequences, so that the model can achieve better disentanglement under the multi-attribute constraints and generate responses with both fidelity to the attributes and relevancy to the context;
3) We design an attribute consistency reward that introduces the direct training signals on control and promote model controllability to attributes.
% }

\section{Our Model}

\subsection{Task Formulation and Model Overview}
Given an input context $\bm{x} = (x_1, ..., x_N)$, our model aims to generate a response $\bm{y} = (y_1, ..., y_M)$ that also satisfies the control attributes $\bm{z}$.
During training, the model jointly learns \textit{response generation} and \textit{attribute prediction}. By doing so, our model is able to generate responses in both scenarios whether control attributes are explicitly given by users or automatically inferred from contexts. We will specify the details later.  

As shown in Figure~\ref{fig:overall_structure}, our model is based on an encoder-decoder framework, which mainly consists of three components:
% Our model is based on an encoder-decoder structure, which is shown in Figure~\ref{fig:overall_structure}. 
1) a \textbf{context encoder} (\S~\ref{subsec:encoder}) first converts an input context into a sequence of hidden states; 2) an \textbf{attribute predictor} (\S~\ref{subsec:attribute_predictor}) predicts the control attributes $\bm{z}$ according to the input context; 3) a \textbf{two-stage controlled decoder} (\S~\ref{subsec:decoder}) takes as inputs the encoder outputs and attributes, and generates a final response in a controllable manner.

\begin{figure*}[t]
    \centering
    \includegraphics[scale=0.8]{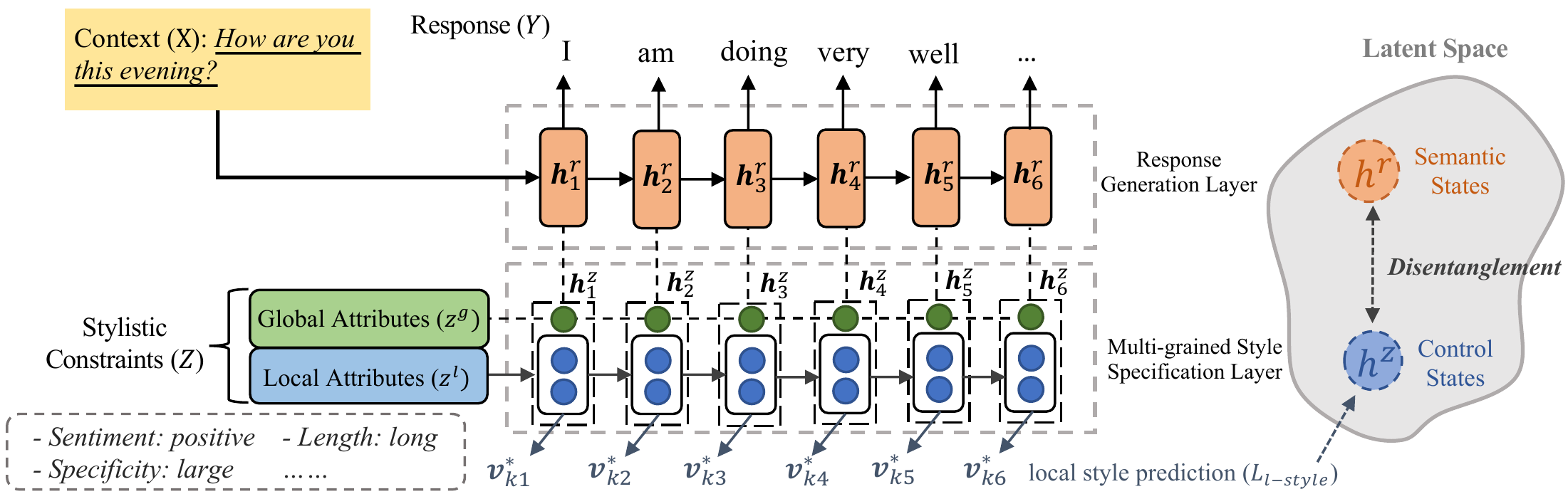}
    \captionof{figure}{ 
    Our two-stage controlled decoder. The \textit{multi-grained style specification layer}
    % (\S~\ref{subsec:control_decoder})
    first generates a sequence of control states $\{\bm{h}_t^z\}$ based on the attributes, and meanwhile predicts local attribute values $\{\bm{v}_{kt}^*\}$ for each local attribute (Eq.~\ref{eq:local}). The \textit{response generation layer}
    % (\S~\ref{subsec:response_decoder})
    then produces the final response based on the control states and context.
    }
    % \vspace{-2mm}
    \label{fig:decoder}
\end{figure*}

\subsection{Context Encoder}
\label{subsec:encoder}
The context encoder maps input contexts into hidden representations using a bi-directional GRU network~\cite{cho2014properties}. Formally, given an input $\bm{x}$, the encoder generates a sequence of hidden states $\{\bm{h}_i^x\}^N_{i=1}$, which will be used for both attribute predictor and initial states of decoder layers.

\subsection{Attribute Predictor}
\label{subsec:attribute_predictor}
We design an attribute predictor to predict each attribute given input context. This benefits our model to generate responses with proper attributes when they are not provided. Particularly, inspired by ~\cite{lian2019learning}, we train the predictor by reducing the prediction divergence between its prior distribution and posterior distribution. 

\smallskip
\noindent \textbf{Prior Distribution.} 
Taking the last context hidden state $\bm{h}_N^x$ as input,
we define the prior distribution as follows:

\vspace{-2mm}
{\fontsize{10}{11}\selectfont
\begin{align}
&{P({z_j|\mathbf x)} =
\mathrm{softmax}(\mathrm{MLP_j^{prior}}(\bm{h}_N^x)),
}
\label{eq:prior}
\end{align}
\vspace{-2mm}
}

\noindent where $z_j$ is the $j$-th attribute, and $\mathrm{MLP_j^{prior}}(*)$ denotes a multi-layer feed-forward network for $z_j$.

\smallskip
\noindent \textbf{Posterior Attribute Distribution.}
Unlike the prior distribution solely based on input context,
the calculation of the posterior distribution involves both input context and response.
Specifically, we first use the same context encoder to learn the semantic representation of the response, forming the response hidden states $\{\bm{h}_t^y\}^M_{t=1}$.
Then, we define the posterior distribution  $P^{\prime}({z_j|\mathbf x, \mathbf y)}$ as follows:

\vspace{-2mm}
{\fontsize{10}{11}\selectfont
\begin{align}
&{P^{\prime}({z_j|\mathbf x, \mathbf y)} =
\mathrm{softmax}(\mathrm{MLP_j^{post}}([\bm{h}_N^x; \bm{h}_M^y])),
}
\end{align}
\vspace{-2mm}
}

\noindent where [;] represents concatenation. 
Compared with the prior distribution, the posterior one better fits the true distributions since responses can also be utilized.

During training, we use gold control attributes with a probability of 80\%, and use predicted attributes from the predictor with a probability of 20\% as the inputs for the decoder 
% \textcolor{blue}{ ## annotation: 2022-0511 ##
to make the model more robust to the inaccurately predicted attributes in test time.  This simple strategy helps to reduce
the attribute discrepancy between model training and test, since only input contexts are available during test time. To generate a response during inference, the attributes can be either provided directly to the model or inferred from prior distributions.
% }  
As discrete control attributes are non-differentiable for gradient backpropagation, we apply the \textit{Gumbel-Softmax Reparameterization} trick~\cite{jang2016categorical} to sample control attributes. 

\smallskip
\noindent \textbf{Training Objective.}
To effectively train the attribute predictor, we define a comprehensive training objective including the attribute prediction loss $(\mathcal{L}_{\text{acc}})$ and the prediction divergence loss $(\mathcal{L}_{\text{kl}})$:

\vspace{-2mm}
{\fontsize{10}{11}\selectfont
\begin{align}
&{\mathcal{L}_{\text{attr}} = \lambda_1{\mathcal{L}_{\text{acc}}} + 
\lambda_2{\mathcal{L}_{\text{kl}}},
}
\end{align}
\vspace{-2mm}
}

\noindent where $\lambda_{*}$ are hyper-parameters. Specifically, $\mathcal{L}_{\text{acc}}$ is used to directly train the predictor, which is defined as:

\vspace{-2mm}
{\fontsize{10}{11}\selectfont
\begin{align}
&{
\mathcal{L}_{\text{acc}} = 
- \log \sum_{j}P^{\prime}(z_j|\mathbf x, \mathbf y)).
}
\end{align}
\vspace{-2mm}
}

Besides, $\mathcal{L}_{kl}$ is approximated as the Kullback-Leibler divergence between the posterior distribution and the prior one:
% We apply Kullback–Leibler Divergence~\cite{kullback1951information} (\textbf{KL-Div}) to approximate the posterior distributions using the prior distributions and minimize the distance of the two:

\vspace{-2mm}
{\fontsize{10}{11}\selectfont
\begin{align}
&{\mathcal{L}_{\text{kl}} =
\sum_j{
\text{KL}(P^{\prime}(z_j|\mathbf x, \mathbf y), P(z_j|\mathbf x)),
}
}
\label{eq:kl}
\end{align}
\vspace{-2mm}
}

\noindent where we use $\text{KL}(*)$ to denote the Kullback-Leibler divergence function. 
% \textcolor{blue}{ ## annotation: 2022-0511 ##
By minimizing $\mathcal{L}_{kl}$, the model learns to enforce the prior and posterior distributions to be as close as possible. 
Note that the attribute prediction loss is beneficial for learning the posterior distribution of the attribute predictor, which in turn helps the prior distribution to approach the same distribution. Thus, our model can sample attributes to generate desirable responses even without input attributes. During training, we set $\lambda_1$ as 1.0 and $\lambda_2$ as 0 in the first 1,000 steps, and then change $\lambda_2$ to 1.0. This enforces the model to firstly learn a good posterior attribute distribution, and then approximate the posterior distribution using the prior distribution.
% }

\subsection{Two-stage Controlled Decoder}
\label{subsec:decoder}
Figure \ref{fig:decoder} shows the basic architecture of our two-stage controlled decoder. We first use a multi-grained style specification layer to generate a sequence of word-level control states based on the control attributes.
Then, we stack a response generation layer to produce  final response with both input context and control states. 
By separating the control states and semantic states, our model is able to address the complicated disentanglement and generate responses maintaining both semantic relevancy to the contexts and fidelity to the attributes.

\subsubsection{\textbf{Multi-grained Style Specification Layer}}
\label{subsec:control_decoder}
We design a novel multi-grained style specification layer to disentangle the \textit{stylistic information} of the response based on the given attributes. 
With attributes $\bm{z}$ as input,
we first introduce an attribute embedding layer to obtain their embeddings. Then, we concatenate all local attribute embeddings and all global ones into 
two control vectors: $\bm{e}^{zl}$ and $\bm{e}^{zg}$, respectively.
Based on these two vectors, 
we finally generate the control state for each response word.

Specifically, given the local control vector $\bm{e}^{zl}$, we first use a GRU network to calculate a sequence of local control states $\{\bm{h}^{zl}_t\}$:

{\fontsize{10}{11}\selectfont
\vspace{-2mm}
\begin{align}
& \bm{h}^{zl}_t = \textsc{GRU}(\bm{h}^{zl}_{t-1}, \bm{k}_t), \\
& \bm{k}_t = \bm{e}^{zl} \odot 
\sigma(\text{MLP}([\bm{h}^{zl}_{t-1}; \bm{e}^{zl}])),
\end{align}
\vspace{-2mm}
}

\noindent where $\sigma(*)$ is a sigmoid nonlinear transformation and $\odot$ represents element-wise multiplication to dynamically inject attributes into each token. Then,
we concatenate each local control state with the global control vector $\bm{e}^{zg}$ to form the final word-level control state, i.e., $\bm{h}^{z}_t = [\bm{h}_t^{zl}; \bm{e}^{zg}]$.

Through the above operations, each  control state is governed by both local and global control vectors, $\bm{e}^{zl}$ and $\bm{e}^{zg}$, where $\bm{e}^{zl}$ is dynamically imposed at each time step,
while $\bm{e}^{zg}$ statically impacts the control state from a global perspective.

Particularly, to enhance the representations of the control states, we further introduce an auxiliary task of local style prediction to predict local attribute values for each response token.~\footnote{
% \textcolor{blue}{ ## annotation: 2022-0511 ##
Different with \cite{ke-etal-2018-generating,wang-etal-2018-learning-ask} predicting the type of word to adjust output distribution, we design this auxiliary task to learn control states that properly reflect the attributes, so that achieve better disentanglement.
% }
} We define the local style prediction loss as

{\fontsize{10}{11}\selectfont
\vspace{-2mm}
\begin{align}
& \mathcal{L}_{\text{l-style}} = 
- \sum_{k=1}^{|\bm{Z}^l|}{\sum_{t=1}^{M}{\log P(\bm{v}_{kt}^{*}|\bm{h}_t^{zl})}}, \label{eq:local}
\end{align}
% \vspace{-1mm}
}

\noindent where $|\bm{Z}^l|$ is the number of local attributes, and $\bm{v}_{kt}^{*}$ represents the ground-truth value of the $k$-th local attribute for the $t$-th response word. 
We will introduce the label construction of $\bm{v}_{kt}^{*}$ in Section
~\ref{construct}.

\subsubsection{\textbf{Response Generation Layer}}
\label{subsec:response_decoder}
On the top of the multi-grained style specification layer, we adopt a response generation layer based on another GRU network to handle the \textit{semantic requirements} and generate the final response.

Concretely, at the $t$-th step, the response generation layer consumes control state $\bm{h}^z_t$ and the previous generated token $y_{t-1}$ to calculate the semantic hidden state $\bm{h}_t^r$:

{\fontsize{10}{11}\selectfont
\vspace{-2mm}
\begin{align}
& \bm{h}^{r}_t = \textsc{GRU}(\bm{h}^{r}_{t-1}, \text{tanh}(\bm{W}_w\bm{y}_{t-1} + \bm{W}_z\bm{h}^z_t)
),
\end{align}
\vspace{-2mm}
}

\noindent where $\bm{W}_*$ are trainable parameters. We also leverage the attention mechanism~\cite{bahdanau2014neural} over the input context to compute a context vector $\bm{c}_t$, and then calculate the probability of the next generated word $y_t$:

{\fontsize{10}{11}\selectfont
\begin{align}
 P(y_t|y_{1:t-1}) &= \text{softmax}(\bm{W}_g[\bm{h}^r_t; \bm{c}_t]) + \bm{b}_g), \\
 \bm{c}_t &= \text{ATT}(\bm{H}^x, \bm{h}^r_t),
\end{align}
\vspace{-2mm}
}

\noindent where $\bm{W}_g$ and $\bm{b}_g$ are trainable parameters, and $\text{ATT}(*)$ represents the attention operation.

\section{Model Training}
Our model is first trained with maximum likelihood (ML) objective. Since ML objective does not provide direct supervision on attribute fidelity, we further employ policy-based reinforcement learning (RL) with an \textit{attribute consistency reward} to continuously train our model. By doing so, we expect our model to generate responses with better fidelity to the control attributes.

\subsection{ML Training Objective}
Our ML training objective mainly includes: negative log-likelihood loss ($\mathcal{L}_{\text{nll}}$), local style prediction loss ($\mathcal{L}_{\text{l-style}}$), constrained bag-of-words loss ($\mathcal{L}_{\text{c-bow}}$), and attribute prediction loss ($\mathcal{L}_{\text{attr}}$):

{\fontsize{10}{11}\selectfont
\vspace{-2mm}
\begin{align}
& \mathcal{L}_{\text{ml}} = \mathcal{L}_{\text{nll}} +
\alpha\cdot\mathcal{L}_{\text{l-style}} + \beta\cdot\mathcal{L}_{\text{c-bow}} + \gamma\cdot\mathcal{L}_{\text{attr}},
\end{align}
\vspace{-2mm}
}

\noindent where $\alpha$, $\beta$ and $\gamma$ are balancing coefficients.

Specifically, we adopt the negative log-likelihood loss for response generation:

{\fontsize{10}{11}\selectfont
\vspace{-2mm}
\begin{align}
& \mathcal{L}_{\text{nll}} = -{\sum_{(\bm{x},\bm{y}, \bm{z})\in D}} \log P(\bm{y}|\bm{x}, \bm{z};\theta),
\end{align}
\vspace{-2mm}
}

\noindent where $\bm{y}$, $\bm{x}$ and $\bm{z}$ are response, context and control attributes, and $\theta$ is model parameter set. 

Inspired by \cite{zhao-etal-2017-learning}, we further design a novel constrained bag-of-words loss to improve the model intepretability and controllability:

{\fontsize{10}{11}\selectfont
\vspace{-2mm}
\begin{align}
& \mathcal{L}_{\text{c-bow}} = 
-\mathbf E_{\bm{z} \sim P(\bm{z}|\bm{x})}
\sum_{t=1}^{M}{\log P_b(y_t|\bm{x}, \bm{z})
},
\end{align}
\vspace{-2mm}
}

\noindent where $P_b(*)$ is an MLP with softmax, which transforms $\bm{h}_N^x$ and $\bm{e}_{z}$ to a probability distribution with the dimension same as vocabulary size $V$, and $\bm{e}_{z}$ is the concatenation of all attribute embeddings. The constrained bag-of-words loss discards word orders and facilitates the model to capture the global semantics of the target response. Also, compared with the original BOW loss, the C-BOW loss optimizes the model to ground the control variables with the explicit semantic information corresponding to the attributes, which enhances the model interpretability from probabilistic perspective, as mentioned in ~\cite{NEURIPS2019_5e2b6675}. 

\subsection{RL with Attribute Consistency}
We apply the self-critical policy gradient training algorithm~\cite{rennie2017self} to use discrete metrics as RL rewards:

{\fontsize{10}{11}\selectfont
\vspace{-2mm}
\begin{flalign} \label{eq:rlrloss}
\mathcal{L}_{\text{rl}} = - (\mathbf{R}(\bm{y}^s) - \mathbf{R}(\hat{\bm{y}}))\log{P(\bm{y}^s |\bm{x}, \bm{z};\theta)},
\end{flalign}
\vspace{-2mm}
}

\noindent where $\bm{y}^s$ is a sampled response obtained by sampling words from $P(\bm{y}^s |\bm{x}, \bm{z};\theta)$, and $\hat{\bm{y}}$ is a self-critical baseline yielded by greedily selecting words that maximize the output probability at each time step. $R(*)$ is the reward function.

\noindent\textbf{Reward Function.} To encourage responses to satisfy the control attributes, we design an attribute consistency reward to measure whether the generated responses are fidelity to the control attributes. The reward of each attribute is calculated at sample level.
% ~\footnote{
% % \textcolor{blue}{ ## annotation: 2022-0511 ##
% The reward of each attribute is calculated in sample-level.
% % }
% } 
Specifically, for discrete attributes such as \textit{Question-asking}, we give a reward of 1 if the response conforms the attribute, or give 0 otherwise. 
For continuous attributes\footnote{The terms of “discrete” and “continuous” attributes are for a different categorization scheme, which are determined by how we compute the value of an attribute, as described in Section
~\ref{construct}.} such as \textit{Specificity}, we first quantize the value into discrete bins, and measure the reverse distance between the response bin value $\hat{z}$ and the attribute bin value $z^*$ as the reward: $1-|\hat{z} - z^*|/{(\#-1)}$, where $\#$ is the number of bins. 
We use the reverse distance for continuous attributes so that the closer $\hat{z}$ is to $z^*$, the larger reward the model will receive. 
The final reward is written as $R(\bm{y}) = {\sum_{z_i}{R_i(\bm{y})}}$.

\section{Experimental Setups}
\subsection{Datasets}
\label{construct}
% \subsection{Datasets and Preprocessing}
We conduct experiments on  \textit{Persona-Chat} and \textit{DailyDialog} datasets. Persona-Chat~\cite{zhang-etal-2018-personalizing} is a conversation dataset grounded on personas, where each participant is assigned with a persona profile serving as background knowledge. We prepend persona texts to dialogue history as the input context. DailyDialog~\cite{li-etal-2017-dailydialog} is a multi-turn chit-chat dataset containing conversations about daily life.We follow the preprocessing as \cite{bao-etal-2020-plato}, and the data statistics are summarized in Table~\ref{tab:stats}. We further filter the samples with a reference length shorter than 3.

 \begin{table}[t]
\fontsize{10}{12}\selectfont
  \centering
    \begin{tabular}{cccc}
        \toprule
        {\bf Dataset} & {\bf Train} & {\bf Val.} & {\bf Test} \\
        \midrule
        {Persona-Chat} & 122,343  & 14,602 & 14,056  \\
        {DailyDialog} & 69,107 & 6,458 & 6,128 \\
        \bottomrule
    \end{tabular}
    \vspace{2mm}
    \caption{
    Statistics on the datasets for experiments.
    }
    \label{tab:stats}
    % \vspace{-7mm}
\end{table}

\begin{table*}[t]
% \fontsize{9}{11}\selectfont
\fontsize{9}{12}\selectfont
 \setlength{\tabcolsep}{2.0mm}
  \centering
%  \setlength{\tabcolsep}{1.3mm}
%   \centering
    \begin{tabular}{l ccccc c ccccc}
        \toprule
        & \multicolumn{5}{c}{\textbf{Persona-Chat}} & \phantom{} & \multicolumn{5}{c}{\textbf{DailyDialog}} \\
        \cmidrule{2-6} \cmidrule{8-12}
        & \textbf{PPL.$\downarrow$} & \textbf{BLEU-1$\uparrow$} & \textbf{BLEU-2} &
        \textbf{Dist.1$\uparrow$}  &
        \textbf{Dist.2} &
        \phantom{} &
        \textbf{PPL.$\downarrow$} & \textbf{BLEU-1$\uparrow$} & \textbf{BLEU-2} & 
        \textbf{Dist.1$\uparrow$} & \textbf{Dist.2}  \\
        \midrule
        
         \rowcolor{blue!15}\multicolumn{12}{l}{\textit{Non-controllable Comparisons}} \\
        \quad\textsc{Seq2seq} & 30.83 & 19.95 & 3.26 & 1.63 & 13.34 &\phantom{} & 30.08 & 19.59 & 2.07 & 3.55 & 23.09 \\
         \quad\textsc{Transformer} & 32.08 & 18.34 & 2.44 & 1.57 & 11.78 &\phantom{} & 29.44 & 18.22 & 1.92 &  4.10 & 22.81 \\
         \quad\textsc{CVAE}$^\ast$ & $\dagger$ & 16.67 & 1.97 & 2.07 & 15.39 &\phantom{} & 39.68 & 14.79 & 1.07 &  4.05 & 26.52 \\
         \quad\textsc{Per-CVAE}$^\ast$ & 40.91 & 17.14 & 1.97 &  \textbf{2.74} & 22.35 &\phantom{} & - & - & - & - & - \\
        % \midrule
        
         \rowcolor{blue!15}\multicolumn{12}{l}{\textit{System Setting (the attributes are not provided and need to be predicted)}} \\
        \quad\textsc{CT-append} & 31.66 & 19.10 & 3.03  & 1.72 & 15.18 &\phantom{} & 34.18 & 19.06 & 1.94 & 3.85 & 25.69 \\
        \quad\textsc{CT-emb} & 33.53 & 19.57 & 3.19 &  1.68 & 13.98 &\phantom{} & 35.79 & 18.40 & 2.04 & 3.82 & 26.48 \\
        \quad\textsc{GTMNES2S}$^\ast$ & 31.36 & 18.84 & 3.24 & 1.03 & 14.66 &\phantom{} & 29.05 & 17.48 & 2.01 & 2.91 & 22.79 \\
        \quad\textsc{CRAYON} (ours) & 27.80 & \textbf{21.07} & \textbf{3.55} & 1.77 & 16.15 &\phantom{} & \textbf{26.80} & \textbf{21.91} & 2.74 & 4.32 & 27.23 \\
        
        \quad\textsc{CRAYON} + RL & \textbf{27.77} & 20.64 & 3.40 & \textbf{2.05} & \textbf{18.49} & \phantom{} & 27.50 & 21.66 & \textbf{2.84} & \textbf{5.02} & \textbf{31.33} \\

        \midrule
        
        \rowcolor{gray!20}\multicolumn{12}{l}{\textit{Oracle Setting (the true attributes are provided)}}  \\
        \quad\textsc{CT-append} & 25.26 & 23.81 & 4.55 & 1.87 & 17.10 &\phantom{} & 26.80 & 23.20 & 2.82 & 3.86 & 26.41 \\
        \quad\textsc{CT-emb} & 26.08 & 23.85 & 4.72 & 2.06 & 17.94 &\phantom{} & 26.94 & 22.28 & 2.74 & 3.92 & 27.55 \\
        \quad\textsc{GTMNES2S}$^\ast$ & 27.79 & 20.44 & 3.77 & 1.06 & 14.68 &\phantom{} & 26.83 & 19.97 & 2.48 & 2.25 & 19.36 \\
        \quad\textsc{CRAYON} (ours) & 21.97 & 25.87 & 5.34 & 2.20 & 20.85 & \phantom{} & \textbf{21.52} & \textbf{25.29} & 3.73 & 4.86 & 32.24 \\
        \quad \textsc{CRAYON} + RL & \textbf{21.87} & \textbf{26.00} & \textbf{5.50} & \textbf{2.46} & \textbf{22.83} & \phantom{} & 21.87 & 25.01 & \textbf{3.86} & \textbf{5.55} & \textbf{35.58} \\
        \bottomrule
        
    \end{tabular}
    \vspace{2mm}
    \caption{%\fontsize{8}{9}\selectfont  
    Experimental results on Persona-Chat and DailyDialog datasets. 
    The best scores are in bold. $^{\dagger}$: perplexity is very unstable due to the sampling process. $^\ast$: results obtained by running code released by
    their authors or our implementation. Our best model variants are significantly better than all comparisons (p $<$ 0.01, Welch’s $t$-test) on perplexity and BLEU, except for BLEU-2 under the system setting on Persona-Chat. The \textit{{Non-controllable}} and \textit{{System Setting}} are comparable where only contexts are available during test time.
  }
%   \vspace{-7mm}
  \label{tab:auto-eval}
\end{table*}

We exploit five control attributes in our experiments: 1) \textbf{Specificity (Spe.)}: following \cite{zhang-etal-2018-learning-control} and \cite{see-etal-2019-makes}, we calculate the specificity value based on the normalized inverse response frequency (NIDF). As NIDF score is continuous, we discretize it into 3 bins; 2) \textbf{Sentiment (Sent.)}: we run Stanford CoreNLP~\cite{manning-etal-2014-stanford} to annotate sentiment results, which can be labeled as ``\textit{positive}'', ``\textit{neutral}'' or ``\textit{negative}''; 3) \textbf{Response-relatedness (Rel.)}: following \cite{see-etal-2019-makes}, we compute response-relatedness based on the cosine similarity between the embeddings of response and the last utterance, and discretize this contineous value into 3 bins; 4) \textbf{Question-asking (Q-A)}: we also consider question-asking as implemented in ~\cite{see-etal-2019-makes}. This binary feature is set to ``\textit{True}'' if and only if at least one word in \{\textit{how, what, when, where, which, who, whom, whose, why, ?}\} appear in the response;
and 5) \textbf{Length (Len.)}: we quantize the response length into 3 bins to represent different size ranges~\cite{fan-etal-2018-controllable}.
% Furthermore, we categorize \textit{Sentiment, Length} and \textit{Question-asking} as global attributes, and \textit{Specificity} and \textit{Response-relatedness} as local attributes based on their properties. 

We categorize \textit{Sentiment, Length} and \textit{Question-asking} as global attributes, and \textit{Specificity} and \textit{Response-relatedness} as local attributes.  
% \textcolor{blue}{ ## annotation: 2022-0511 ##
The categorization is based on the property whether the attribute can be directly reflected on each token or it only affects the generation from an overall perspective.
% }
In particular, we consider \textit{Specificity} as a local attribute since the specificity of the whole response strongly depends on the specificity score of each word. Besides, we consider \textit{Response-relatedness} as a local attribute because 1) the relatedness of response and context can be reflected on each response token (the sentence embeddings are calculated as the weighted sum of word embeddings) and 2) \cite{see-etal-2019-makes} find that weighted decoding is more effective for response-relatedness, where they consider the similarity between each response token and context as the decoding feature. Thus, we incorporate the token-level relatedness into local attribute prediction to bring a performance gain.

For local planning loss as in Equation~\ref{eq:local}, we construct local attribute labels for each response word. For \textit{Specificity}, we discretize the NIDF score of each token into 6 bins. For \textit{Response-relatedness}, we compute the cosine similarity between the word embedding and the embedding of the last utterance, and then quantize this similarity into 6 bins.

\subsection{Baselines and Settings}
We first compare our model with the following \textit{non-controllable baselines}: 1) \textsc{Seq2seq}~\cite{bahdanau2014neural}: the standard sequence-to-sequence model with attention mechanism; 
% (2) \textbf{Persona-Seq2Seq}: this is the same as Seq2seq except for prepending persona texts to context as input; 
2) \textsc{Transformer}~\cite{vaswani2017attention}: the standard transformer-based model which has been proved effective for text generation tasks; 3) \textsc{CVAE}~\cite{zhao-etal-2017-learning}: a conditional variational autoencoder that captures the discourse-level diversity;
4) \textsc{Per-CVAE}~\cite{song2019exploiting}: a memory-augmented architecture with incorporation of explicit persona texts. 
Besides, we consider  \textit{controllable baselines} including 1) GTMNES2S~\cite{xu-etal-2019-neural}: a sequence-to-sequence architecture with a goal tracking memory network;\footnote{Since the source code of GTMNES2S are not released, we implement their method.
}  2) \textsc{CT}~\cite{see-etal-2019-makes,keskar2019ctrl}: the conditional training method which directly incorporates control attributes into inputs. 
For \textsc{CT}, we report the performance of two variants: \textsc{CT-append} that appends the attributes to the input contexts and \textsc{CT-emb} which concatenates the embeddings of control attributes to the decoder’s input at every step. We apply the same attributes as our model for controllable baselines.

We employ both \textit{system setting} and \textit{oracle setting} for controllable models.
Under the system setting, the attributes are not given and need to be predicted by the model based on contexts, and we focus on evaluating the response quality. Under the oracle setting, the true attributes are provided, and we mainly focus on evaluating the model controllability.

\subsection{Traning Details}
We implement our model using  OpenNMT~\cite{klein-etal-2017-opennmt}. We use a two-layer bidirectional GRU for both encoder and decoder with the 300-dimensional hidden size (150 per direction). For Transformer, we apply a 4-layer encoder and decoder with 6 attention heads and the dimension of 300.
We initialize word embeddings with GloVe~\cite{pennington-etal-2014-glove} and fine-tuned them during training. The dimension of attribute embeddings is 300, which is randomly initialized.
The batch size is 64. We apply the Adam optimization~\cite{kingma2014adam} with warm-up steps of 500 and the maximal learning rate of 5e-4. 
We implement early stopping based on the perplexity on validation set. Also we set dropout with a retention probability of 0.9 to prevent over-fitting. All parameters are tuned from the validation set.
To make the model more robust to attributes under system setting, we apply schedule sampling during training: we use gold control attributes with a probability of 80\%, and use predicted attributes from the predictor with a probability of 20\%.
During the RL stage, since only incorporating RL loss leads to a degradation of response fluency, we apply a combination of RL loss and NLL loss.

For PER-CAVE, we use their implementations.\footnote{\url{https://github.com/vsharecodes/percvae}}
For GTMNES2S, since the source code is not released, we reproduced the model according to their paper. We apply the same attributes as our model, and set the hidden size and number of layers the same as our model.

All experiments are trained on NVIDIA Tesla V100 GPUs.  For our model, it takes approximately 2 hours to converge in the ML training stage, and 10 hours in the RL training stage. 
When selecting the model checkpoint, we choose the one based on the validation perplexity.

\section{Results}
\subsection{Automatic Evaluation}
We adopt perplexity (PPL.) to evaluate response fluency and BLEU-1/2~\cite{papineni-etal-2002-bleu} to evaluate relevance. 
For diversity, we employ Distinct-1/2 (Dist.1/2)~\cite{Li2016naacl} to calculate the ratio of distinct uni-grams or bi-grams. The results are shown in Table~\ref{tab:auto-eval}. 

\smallskip
\noindent\textbf{Comparison with Non-controllable Baselines.}
Under the system setting, where the same inputs (only contexts) are leveraged, our model achieves remarkably lower perplexity than all non-controllable baselines on both datasets. These results indicate that our model is able to generate more fluent responses. As for relevance, our model significantly outperforms the baselines in terms of BLEU-1 and BLEU-2.\footnote{We cannot conduct significant test on Dist.1/2 since we compute the inter-distinct on total generated words as implemented in ~\cite{Li2016naacl}.}
Furthermore, our RL variant produces more diverse responses with larger distinct-1/2. Although \textsc{Per-CAVE} achieves high distinct scores, its perplexity and BLEU scores are very low, indicating a low response quality. By contrast, our model can achieve both high response quality and diversity. All these improvements demonstrate that by incorporating important control attributes, our model can produce more appropriate and diverse responses. 

\begin{figure}[t]
    \centering
    \includegraphics[scale=0.32]{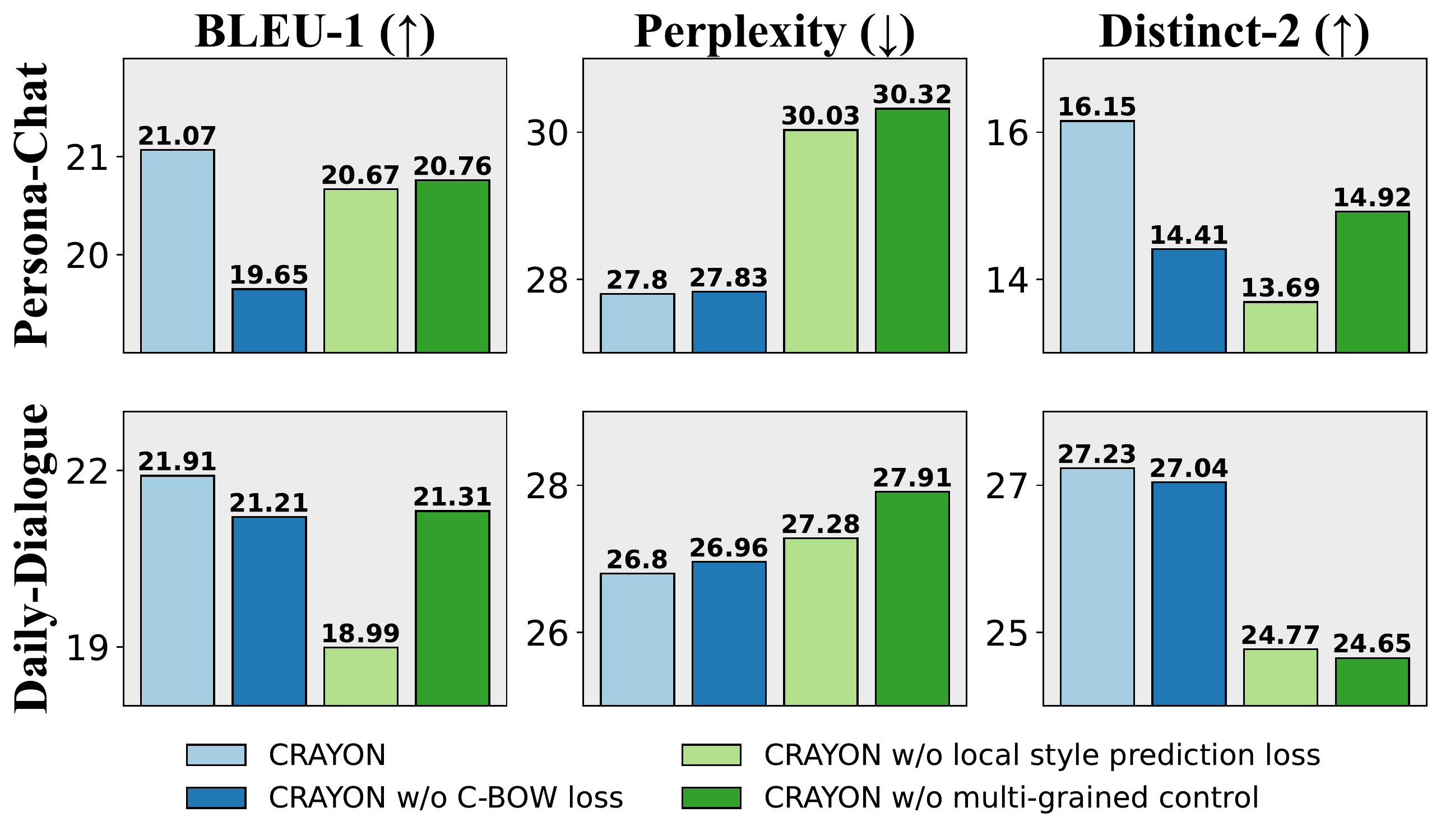}
    \vspace{-1mm}
    \captionof{figure}{ 
    Ablation results of our model variants.
    }
    \label{fig:ablation}
    % \vspace{-4mm}
\end{figure}

\smallskip
\noindent\textbf{Comparison with Controllable Baselines.}
Moreover, compared to CT and GTMNES2S that directly incorporate attributes into input contexts and rely on one single module to deal with both stylistic constraints and semantic requirements altogether, our model surpasses these baselines towards all aspects under both system and oracle settings. 
The results prove the effectiveness of our model on disentangling the complicated controllable generation under multi-attribute constraints.
Furthermore, after applying RL training, the distinct scores are improved, showing that introducing explicit training signals on attributes benefits the model to generate more diverse responses. This is consistent with our motivation to tackle the one-to-many mapping problem via explicit control. Overall, the above results demonstrate that our model can achieve better response quality and diversity than controllable baselines.

\smallskip
\noindent\textbf{Ablation Study on Model Variants.}
We further analyze our model variants to quantify the contributions of various components under the system setting. As shown in Figure~\ref{fig:ablation}, \textsc{CRAYON} achieves the best performance towards both response quality and diversity. Meanwhile, removing C-BOW loss ($\mathcal{L}_{\text{c-bow}}$) and local style prediction loss ($\mathcal{L}_{\text{l-style}}$) leads to performance degradation. In particular, removing local style prediction loss brings significant decreases to response diversity. The results imply the effectiveness of these two losses. Finally, we consider the variant without multi-grained control, where we treat global attributes as local ones with identical labels for all tokens, and the results also 
drop. This result indicates that incorporating attributes in different granularities indeed help the model to effectively steer the generation process.

\subsection{Human Evaluation}
% \noindent\textbf{Human Evaluation.}
\begin{table}[t]
\fontsize{9}{11}\selectfont
    \def\arraystretch{0.7}
    \centering
    \begin{tabular}{lp{70mm}}
         \toprule
         \multicolumn{2}{c}{\textbf{Readability:}} \\
         \midrule
         \rowcolor{lightgray!30}
         1 & Not readable, contains fragments, missing components, or serious grammar errors  \\
         3 & Contains relatively minor grammatical errors, not very fluent, but understandable \\
         \rowcolor{lightgray!30}
         5 &  Correct Grammar, very fluent and complete \\
         \midrule
         \multicolumn{2}{c}{\textbf{Coherence:}} \\
         \midrule
         \rowcolor{lightgray!30}
         1 & Not relevant with the contexts, or inconsistent with dialogue history or background knowledge \\
         3 & Relevant to the contexts, but with minor conflicts to the dialogue history or background knowledge \\
         \rowcolor{lightgray!30}
         5 & Completely coherent and relevant to the dialogue contexts and background knowledge. \\
         \midrule
         \multicolumn{2}{c}{\textbf{Content richness:}} \\
         \midrule
         \rowcolor{lightgray!30}
         1 & Very generic or boring. Do not want to continue this conversation \\
         3 & Contains some information, but somewhat not interesting or informative \\
         \rowcolor{lightgray!30}
         5 & Interesting, informative and you want to continue the conversation. \\
         \midrule
         \multicolumn{2}{c}{\textbf{Overall quality:}} \\
         \midrule
         \rowcolor{lightgray!30}
         1 & Not a valid response \\
         3 & Can be a response, but contains some language errors or not informative \\
         \rowcolor{lightgray!30}
         5 & A good response. \\
         \bottomrule
         \vspace{-1mm}
    \end{tabular}
    \caption{Explanations on human evaluation aspect scales.}
    \label{tab:human_eval}
\end{table}

\begin{table}[t]
\fontsize{8}{11}\selectfont
  \centering
    \begin{tabular}{l ll l c}
        \toprule
        & \textbf{Read.} & \textbf{Coh.} & \textbf{Rich.} & \textbf{Overall.} \\
        \midrule
        \textsc{Seq2seq} &  3.13 & 3.10 & 2.85 & 3.14 \\
        \textsc{Per-CVAE} &  2.74 & 2.54 & 2.62 & 2.77 \\
        \textsc{CT-append} & 2.76 & 2.59 & 2.69 & 2.81 \\
        \textsc{CT-emb} & 2.59 & 2.63 & 2.57 & 2.59 \\
        \textsc{GTMNES2S} & 2.74 & 2.38 & 3.13 & 2.77 \\
        % \textsc{Transformer} & 2.78 & 1.60 & 2.17 & 2.12 \\
        \textsc{CRAYON} (ours) & \textbf{3.29} & \textbf{3.30}$^\ast$ & 3.20 &
        \textbf{3.41}$^\ast$ \\
        \textsc{CRAYON} + RL & 3.17 & 3.15 & \textbf{3.36}$^\ast$ & {3.38$^\ast$} \\
        \bottomrule
    \end{tabular}
    \vspace{2mm}
    \caption{ %\fontsize{9}{11}\selectfont
    Human evaluation with scores on a scale of 1 to 5 (best). $^\ast$: significantly better than all comparisons (p $<$ 0.005, Welch’s $t$-test). The Krippendorf’s $\alpha$ values for all aspects exceed 0.4, indicating general
    consensus to intermediate agreement.
  }
%   \vspace{-6mm}
  \label{tab:human-eval-quality}
\end{table}

We also conduct human evaluation to analyze the response quality. 
Concretely, we randomly select 100 examples from Persona test set, and generate outputs using both our model variants and the baselines. We hire three proficient English speakers as human annotators, and the annotators are asked to independently evaluate the quality of outputs along with the persona and dialogue history on a Likert scale of 1 (worst) to 5 (best) regarding: (1) {\bf Readability (Read.):} whether the response is fluent, complete, grammatically correct and can be understood; (2) {\bf Coherence (Coh.):} whether the response is relevant with the dialogue context and consistent with the dialogue history or background knowledge, and this is highly correlated with the \textit{Response-relatedness} control attribute; (3) {\bf Content richness (Rich.):} whether the response is informative, interesting and encourages you to continue the conversation; and (4) {\bf Overall quality (Overall.):} this is a general assessment that whether you think it is a good response or not.
To avoid bias, we anonymize the models and shuffle the outputs to the annotators.
More details are in Table~\ref{tab:human_eval}.

As shown in Table~\ref{tab:human-eval-quality}, our model variant achieves the highest scores on all aspects. After applying the attribute consistency reward, the readability and coherence slightly decrease. We find that introducing RL training makes the responses sometimes contain grammatical errors, but meanwhile the model generates more diverse and informative responses with better content richness. Notably, \textsc{CT-emb} and \textsc{CT-append}  produce low results. We hypothesize that directly adding all attributes to inputs brings complex disentanglement issues and the model is hard to learn the correct relation between attributes and responses.

\begin{table}[t]
\fontsize{8}{10}\selectfont
  \centering
    \begin{tabular}{lcc lcc}
        \toprule
        & {\bf Q-A.} & {\bf Len.} & {\bf Sent.} & {\bf Rel.} & {\bf Spe.} \\
        \midrule
        \rowcolor{gray!20}
        \textit{Persona-Chat} &  &  &  &  &  \\
        \quad\textsc{CT-append} & 92.57 & 81.12 & 76.15 & 63.48 & 54.49  \\
        \quad\textsc{CT-emb} & 96.44 & 80.69 & 74.59 & 63.74 & 58.78 \\
        \quad\textsc{GTMNES2S} & 97.15 & 70.16 & 74.90 & 67.58 & 62.43  \\
        \quad\textsc{CRAYON} (ours) & 98.32 & 85.21 & 75.09 & 69.99 & 64.46 \\
        \quad\quad\quad \textit{w/o $\mathcal{L}_{\text{l-style}}$} & 97.52 & 82.10 & 72.30 & 66.19 & 52.61 \\
        \quad\textsc{CRAYON} + RL & \textbf{98.73} & \textbf{85.31} & \textbf{84.28} & \textbf{73.87} & \textbf{75.10}\\
        \midrule
        \rowcolor{gray!20}
        \textit{DailyDialog} & & & & & \\
        \quad\textsc{CT-append} & 92.54  & 76.41 & 72.05 & 71.63 & 52.10  \\
        \quad\textsc{CT-emb} & 91.78 & 81.69 & 75.92 & 73.68 & 56.76 \\
        \quad\textsc{GTMNES2S} & 93.88 & 67.38 & 72.98 & 74.38 & 60.94  \\
        \quad\textsc{CRAYON} (ours) & 96.48  & 88.35 & 76.62 & 76.59 & 58.19\\
        \quad\quad\quad \textit{w/o $\mathcal{L}_{\text{l-style}}$} & 96.33 & 83.56 & 75.77 & 74.76 & 55.65 \\
        \quad\textsc{CRAYON} + RL & \textbf{98.17} & \textbf{90.29}  & \textbf{82.17} & \textbf{79.65} & \textbf{67.40}\\
        \bottomrule
    \end{tabular}
    \vspace{2mm}
    \caption{
    Control Accuracy (\%) with oracle attributes.
  }
  \label{tab:control_acc1}
%   \vspace{-4mm}
\end{table}

\begin{table}[t]
\fontsize{8}{10}\selectfont
  \centering
    \begin{tabular}{lcc lcc}
        \toprule
        & {\bf Q-A.} & {\bf Len.} & {\bf Sent.} & {\bf Rel.} & {\bf Spe.} \\
        \midrule
        \rowcolor{gray!20}
        \textit{Persona-Chat} &  &  &  &  &  \\
        \quad\textsc{CT-append} & 87.62 & 79.90 & 71.35 & 54.64 & 46.79  \\
        \quad\textsc{CT-emb} & 95.03 & 78.15 & 65.82 & 53.60 & 50.28 \\
        \quad\textsc{GTMNES2S} & 89.76 & 35.14 & 69.89 & 56.93 & 41.14  \\
        \quad\textsc{CRAYON} (ours) & 96.64 & 82.04 & 65.44 & 56.67 & 55.85 \\
        \quad\textsc{CRAYON} + RL & \textbf{97.82} & \textbf{82.47} & \textbf{78.23} & \textbf{61.40} & \textbf{65.82}\\
        \midrule
        \rowcolor{gray!20}
        \textit{DailyDialog} & & & & & \\
        \quad\textsc{CT-append} & 86.84 & 71.49 & 60.84 & 47.02 & 46.57  \\
        \quad\textsc{CT-emb} & 85.06 & 76.87 & 61.52 & 48.09 & 60.10 \\
      \quad\textsc{GTMNES2S} & 91.02 & 49.46 & 62.48 & 46.69 & 46.97  \\
        \quad\textsc{CRAYON} (ours) & 92.64 & 84.28 & 59.56 & 49.67 & 51.69\\
        \quad\textsc{CRAYON} + RL & \textbf{94.97} & \textbf{86.49}  & \textbf{68.12} & \textbf{51.12} & \textbf{60.92}\\
        \bottomrule
    \end{tabular}
    \vspace{2mm}
    \caption{
    % \textcolor{blue}{ ## annotation: 2022-0511 ##
    Control Accuracy (\%)  with probing attributes.
    % }
  }
  \label{tab:control_acc2}
%   \vspace{-4mm}
\end{table}

\begin{figure}[t]
    \centering
    \includegraphics[scale=0.32]{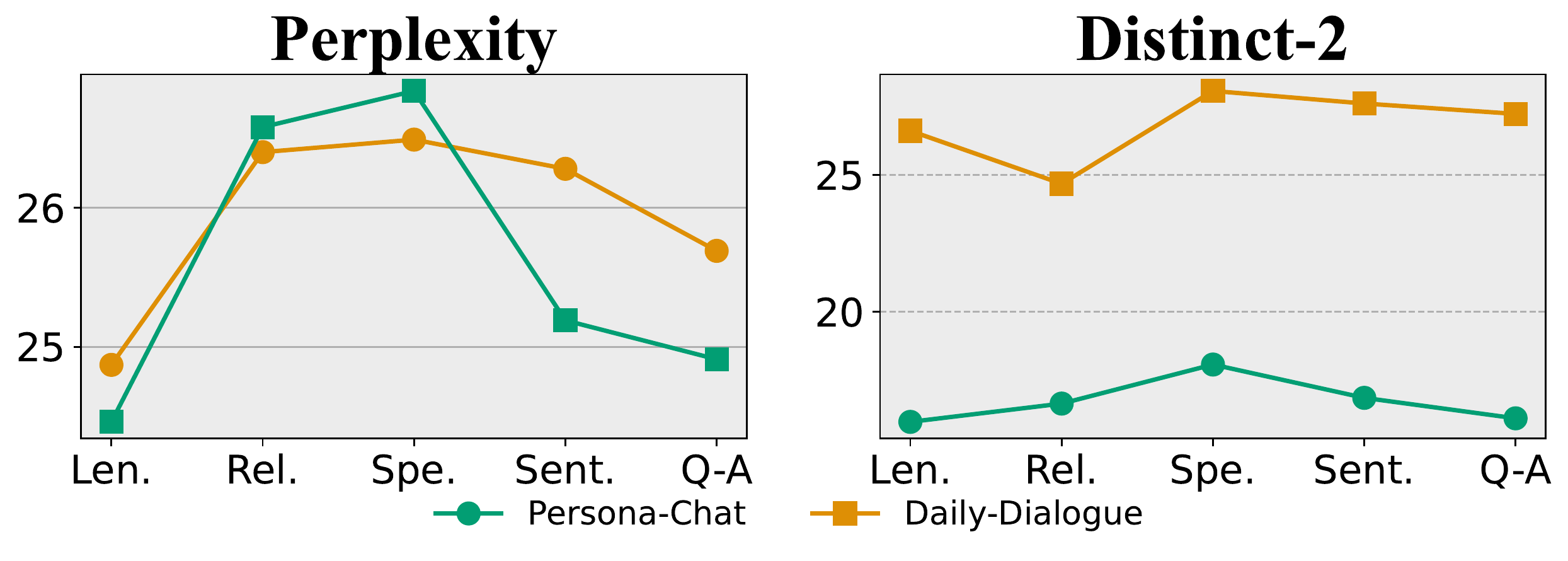}
    \vspace{-3mm}
    \captionof{figure}{ 
    Contributions of each single control attributes.
    }
    \label{fig:single_attributes}
    % \vspace{-4mm}
\end{figure}

\begin{figure*}[t]
    \def\arraystretch{1.8}
    \bgroup
    \def\arraystretch{1.8}
	\fontsize{9}{11}\selectfont
     \hspace{-1mm}
	\setlength{\tabcolsep}{1.0mm}
	\centering
	\begin{tabular}{p{120mm}}
% 	\hline
    \toprule
	\quad\textbf{Context}: 
	
	\quad\textit{A}: How are you this evening?
	
% 	\hline
	\quad\textbf{Response}:
	
	\quad{\color{blue!55} {\footnotesize \circled{1}} Q-A=False, Len.=1, Sent.=Negative, Rel.=1, Spe.=0}
	
	 \quad\textit{Generated}: {\textit{\underline{Tired}}}. Watching a movie on the weekends.
	 
	\quad{\color{blue!55}{\footnotesize \circled{2}} Q-A=False, Len.=1, Sent.=Neutral, Rel.=0, Spe.=0}
	
	\quad\textit{Generated}: \textit{\underline{Not bad}}. Just listening to rap music, cooking some pizza.

	\quad{\color{blue!55}{\footnotesize \circled{3}} Q-A=True, Len.=2, Sent.=Positive, Rel.=0, Spe.=2}
	
	\quad\textit{Generated}: I am \textit{\underline{doing very well}}. Just finished eating \textbf{\textcolor{intro_purple}{sushi}}. Where are you from?
	\\

    \hline
    \quad\textbf{Context}: 
	
	\quad\textit{A}: Hey guess what? I am a jewish boy and still enjoy jazz music.
	
	\quad\textit{B}: Cool! I am jewish too! I play pop music, i am the lead singer of bb2.
	
	\quad\textbf{Response}:
	
	\quad{\color{blue!55} {\footnotesize \circled{1}} Q-A=False, Len.=1, Sent.=Positive, Rel.=1, Spe.=0}
	
	 \quad\textit{Generated}: That is \textit{\underline{pretty cool}}. I love music too!
	 
	\quad{\color{blue!55}{\footnotesize \circled{2}} Q-A=False, Len.=2, Sent.=Positive, Rel.=1, Spe.= 2}
	
	\quad\textit{Generated}: \textit{\underline{Awesome}}! I am a \textbf{\textcolor{intro_purple}{jazz}} singer. I love music and \textbf{\textcolor{intro_purple}{jazz}} music mostly. \\

    % \hline
    \bottomrule
	\end{tabular}
\caption{Sample responses with corresponding attributes, generated by our model with RL on Persona-Chat test set. Words with large specificity scores are in boldface and color-coded.
} 
% \vspace{-1mm}
\label{fig:samples}
\egroup
\end{figure*}

\subsection{Analysis on Model Controllability}
% \smallskip
\noindent\textbf{Accuracy of Control on  Attributes.}
We study the control accuracy of each attribute to measure the model controllability in Table~\ref{tab:control_acc1}. 
Specifically, we compute the attribute values of each generated response with the same classifiers as in data preprocessing and compare the values with the true input attributes. First, our model trained with ML objectives outperforms \textsc{CT} and \textsc{GTMNES2S} on both datasets. This shows that our multi-grained style specification layer can accurately reflect the stylistic constraints and help to produce desired responses that are conformable to the attributes.  Second, removing local style prediction loss leads to performance decreases, indicating the effectiveness of this auxiliary task on strengthening the style specification.
Third, the observation of the normal accuracy on \textit{Specificity} is consistent with prior work~\cite{see-etal-2019-makes}. We speculate that \textit{Specificity} is a relatively implicit attribute, and the small-scale training data is insufficient for the model to learn the mapping between the attribute and response. Nonetheless, the results in Figure~\ref{fig:single_attributes} empirically show that the model can still produce high-quality responses with this attribute.
Fourth, after applying RL with attribute consistency reward, the results of all attributes are improved, especially for \textit{Specificity}, proving that our model can generate responses with better fidelity to the attributes by introducing explicit supervision signals.

% \smallskip
% \noindent\textbf{Accuracy of Control on Attributes.}
% \textcolor{blue}{ ## annotation: 2022-0511 ##
In order to further explore the controllability of the model, 
we conduct probing experiments by trying all possible values of each attribute, and then compute the control accuracy of models. Taking \textit{Length} as an example, we enumerate all possible values of \textit{Length} for each sample while leaving the other attributes unchanged as the ground-truth response. By doing this to all attributes, we obtain multiple probing attribute values.
As shown in Table~\ref{tab:control_acc2}, our model variant achieves the highest scores on all probing attributes. Notably, compared with the accuracy on the oracle attributes in Table~\ref{tab:control_acc1}, the scores on probing attributes declined. 
For this result, we speculate that the underlying reasons include two aspects: 1) the incompatibility between attribute values and context, and 2) the conflicts between probing attribute values.
% }

\smallskip
\noindent\textbf{Ablation on Single Attribute.} 
We study the contribution of each attribute in Figure~\ref{fig:single_attributes}.~\footnote{
% \textcolor{blue}{ ## annotation: 2022-0511 ##
All the attributes that appear in this paragraph refer to the oracle attribute.
% }
} 
We consider one single attribute  to control the response generation at each time, and present the corresponding perplexity and distinct-2.  
As shown in the left figure, among all attributes, \textit{Length} and \textit{Question-asking} are effective on both datasets. 
While \textit{Sentiment} is quite useful on DailyDialog, it does not contribute much on Persona-Chat. In the right figure, we can conclude that \textit{Specificity} is more relevant to the response diversity, while \textit{Length} and \textit{Response-relatedness} are less relevant. This indicates that introducing proper attributes is critical.
In addition, compared with the results in Table~\ref{tab:auto-eval}, incorporating all attributes can significantly improve the model performance towards both response fluency and diversity, further proving the importance of introducing controls into dialogue generation.

\begin{table}[t]
\fontsize{8}{10}\selectfont
 \setlength{\tabcolsep}{1.3mm}
  \centering
    \begin{tabular}{lccccc}
        \toprule
        & {\bf PPL.} & {\bf BLEU-1} & {\bf BLEU-2} & {\bf Dist.1} & {\bf Dist.2} \\
        \midrule
        % \quad\textsc{Finetune} & 5.890 & 22.35 & 4.85 & 2.06 & 15.86 \\
        \quad\textsc{CT-append (w/ bart)} & 5.58 & 27.28 & 6.70 & 1.78 & 14.95 \\
        \quad\textsc{CRAYON (w/ bart)} & \textbf{4.76} & \textbf{28.14} & \textbf{7.05} & \textbf{2.20} & \textbf{18.54} \\
        \bottomrule
    \end{tabular}
    \vspace{2mm}
    \caption{
    % \textcolor{blue}{ ## annotation: 2022-0511 ##
    Automatic results on Persona-Chat with BART.
    % }
  }
  \label{tab:pretrain_eval}
  \vspace{-2mm}
\end{table}

\begin{table}[t]
\fontsize{8}{10}\selectfont
  \centering
    \begin{tabular}{lccccc}
        \toprule
        & {\bf Q-A.} & {\bf Len.} & {\bf Sent.} & {\bf Rel.} & {\bf Spe.} \\
        \midrule
        % \quad\textsc{Finetune} & 57.56 & 40.32 & 48.25 & 39.63 & 40.48  \\
        \quad\textsc{CT-append (w/ bart)} & 97.50 & 90.38 & 68.66 & 62.64 & 53.59 \\
        \quad\textsc{CRAYON (w/ bart)} & \textbf{99.29} & \textbf{94.24} & \textbf{73.58} & \textbf{64.77} & \textbf{55.26}  \\
        \bottomrule
    \end{tabular}
    \vspace{2mm}
    \caption{
    % \textcolor{blue}{ ## annotation: 2022-0511 ##
    Control Accuracy (\%) on Persona-Chat with BART.
    % }
  }
  \label{tab:pretrain_control}
  \vspace{-2mm}
\end{table}

\subsection{Incorporation of Pretrained Models}

Here we discuss the adaptation ability of our method to pretrained language models. We adopt BART~\cite{lewis-etal-2020-bart} as the base model due to its promising results on many text generation tasks. Specifically, for \textsc{CT-append}, we directly append the control attributes to the context as the input, which are then fed to BART for response generation. For \textsc{CRAYON}, we combine the control states from  style specification layer and the last decoder hidden state of BART to generate the response. The experiments are conducted on the Person-Chat data under oracle setting, where true attributes are provided in both training and inference without attribute prediction. To better discern the effectiveness of our disentangled style specification layer, we further exclude the RL training stage.

The automatic results are shown in Table~\ref{tab:pretrain_eval}. Our proposed model outperforms \textsc{CT-append (w/ bart)} on all aspects, demonstrating the effectiveness of our method to be incorporated into pretrained models. We further compute the control accuracy, and the results are presented in Table~\ref{tab:pretrain_control}. As we can see, our model variant with BART achieves better controllability on all attributes  compared with \textsc{CT-append}. This further indicates our style specification layer is useful to address the disentanglement of stylistic constraints and generate responses with better fidelity to the attributes. The above results prove the generalizability of our proposed method, which can be flexibly incorporated into pretrained models for better results.

\begin{figure*}[t]
    \def\arraystretch{1.5}
	\fontsize{9}{11}\selectfont
        \hspace{-1mm}
	\setlength{\tabcolsep}{1.0mm}
	\centering
	
	\begin{tabular}{p{120mm}}
	\bottomrule
        % \hline
        \begin{tabular}{p{55mm}|p{65mm}}
        \textbf{Persona} & \textbf{Context} \\
        \makecell[l]{
         \tabitem I do not drive because I live in new \\\quad\, york \\
         \tabitem I like jazz music \\
         \tabitem My favorite tv show is the office \\
         \tabitem I do not eat anything sweet \\
        } & 
        \makecell[l]{
        \break
        (......) \\
        \textbf{A}: I try to not eat many sweets to set a 
        good \\ example for my boys. \\
        \textbf{B}: I do not like them that much. Do you 
        have \\ any hobbies? \\
        \textbf{A}: I am single mom so other than hiking, a \\ 
        lot of running my boys around to sports
        } \\
        \end{tabular} \\
        \toprule
        % \hline
        \textbf{\textsc{Per-CVAE: }}   {I' m a bit of a vegan. I can not wait for that.} \\
        
        \hline
        \textbf{\textsc{Seq2Seq: }}  {I enjoy running too. } \\
        
        \hline
        \textbf{\textsc{CT-append: }}  {That is good. I'm on road trips. } \\
        
        \hline
        \textbf{\textsc{CT-emb: }}  {Oh cool. I love to hike. I have only 2 kids, but i'm a UNK too!} \\
        
        \hline
        \textbf{\textsc{GTMNES2S: }}  {That is good. I have long hair and dyed my hair purple!} \\
        
        \hline
        \textbf{\textsc{Transformer: }}  {It is a great place to go}\\
        
        \hline
        \textbf{\textsc{CRAYON: }}  {That is fun. I like to learn about history and watching TV. } \\
        
        \hline
        \textbf{\textsc{CRAYON + RL: }}  {That's a great hobby, I also enjoy listening to music while playing}\\
        % \hline
        \bottomrule
    \end{tabular}
    \vspace{2mm}
\caption{
Sample outputs of our model variants and baselines on Persona-Chat dataset under the system setting. For controllable methods, the oracle attribute values are not given and need to be predicted.
}
\label{tab:persona-sample-5}
\vspace{-2mm}
\end{figure*}

\subsection{Case Study}

We show sample outputs of our model with the corresponding attributes in
Figure~\ref{fig:samples}. Our model is able to generate proper responses with desired
attributes. Specifically, given a negative sentiment, our model correctly generates \textit{``tired"}, while for the positive sentiment our model generates phrases such as \textit{``very well"} and \textit{``pretty cool"}. Besides, when a high specificity is given, our model produces response words such as \textit{``sushi"}, \textit{``jazz"} and \textit{``romance"}. 
With respect to \textit{Question-asking}, the model learns to correctly use \textit{``?''}. Overall, our model can properly reflect the stylistic constraints and generate responses conformable to the controls. This proves the effectiveness of our model regarding controllability by incorporating fine-grained controls into the two-stage controlled decoder.

Figure~\ref{tab:persona-sample-5} shows sample outputs generated by our model variants and baselines under the system setting. Compared to baseline outputs, our generated response is more coherent to the context and consistent to the persona texts. In addition, our RL variant generates a more diverse response with specific talking points such as \textit{``listening to music"}.
% More samples are provided in the appendix material.

\section{Conclusion}
In this paper, we have presented \textsc{CRAYON}, a controllable dialogue generation model with a novel two-stage controlled decoder and an attribute consistency reward to steer the generation process in a way of multi-grained controls. Both automatic and human evaluations indicate that our proposed model can address the complicated disentanglement and generate high-quality responses with better fidelity to controls.
In the future, we plan to extend our method to other generation tasks.

\ifCLASSOPTIONcaptionsoff
  \newpage
\fi

\bibliographystyle{IEEEtran}
\bibliography{taslp22}

\end{document}